\documentclass[11pt]{article}

\usepackage[final]{acl}

\usepackage{times}
\usepackage{latexsym}

\usepackage[T1]{fontenc}

\usepackage[utf8]{inputenc}

\usepackage{microtype}

\usepackage{listings}
\definecolor{grewkeyword}{RGB}{0,70,140}   
\definecolor{grewattr}{RGB}{153,0,153}     
\definecolor{grewstring}{RGB}{200,0,0}     
\definecolor{grewcomment}{RGB}{0,128,0}    

\lstdefinelanguage{GREW}{
    alsoletter={:},
    morekeywords={
        upos,Case,Gender,Number,Animacy,AdvType,Person,POS,
        obj,advmod:neg,case,obl,nsubj,root
    },
    sensitive=true,
    morestring=[b]",
    morestring=[b]',
    morecomment=[l]{//},
}

\lstdefinestyle{grewstyle}{
    language=GREW,
    basicstyle=\ttfamily\small,
    keywordstyle=\color{grewkeyword}\bfseries,
    stringstyle=\color{grewstring},
    commentstyle=\color{grewcomment}\itshape,
    numbers=none,
    breaklines=true,
    showstringspaces=false,
    columns=fullflexible,
    literate={->[}{{\textcolor{grewattr}{->}}}2
             {:[}{{\textcolor{grewattr}{:}}}1
             {ö}{{\"o}}1 {ä}{{\"a}}1 {ü}{{\"u}}1 {ß}{{\ss}}1
             {Ö}{{\"O}}1 {Ä}{{\"A}}1 {Ü}{{\"U}}1
}

\usepackage{inconsolata}

\usepackage{graphicx}

\usepackage{polyglossia}

\usepackage[utf8]{inputenc}
\usepackage{verbatim}
\usepackage[T1]{fontenc}
\usepackage{textcomp}  
\usepackage{subcaption,graphicx}
\usepackage{mathtools}  
\usepackage{amssymb}    
\usepackage{amsthm}

\usepackage{amssymb}
\usepackage{expex}
\usepackage{tipa} 
\usepackage{multirow}
\usepackage[table, dvipsnames]{xcolor}
\definecolor{softgreen}{RGB}{100,230,100}

\usepackage{booktabs}
\usepackage[table]{xcolor}
\usepackage{adjustbox}

\usepackage{listings}

\usepackage{booktabs}
\usepackage{makecell} 
\usepackage{graphicx} 
\usepackage{rotating}

\usepackage{siunitx}
\usepackage{subcaption} 
\usepackage{pifont}
\usepackage{array}
\usepackage{pifont}
\usepackage{subcaption}
\usepackage{amsmath}
\usepackage{longtable} 

\title{Targeted Syntactic Evaluation of Language Models on \newline Georgian Case Alignment}

\author{Daniel Gallagher \and Gerhard Heyer \\
  Institute for Applied Informatics (InfAI), Leipzig \\
  \texttt{gallagher@infai.org, heyer@infai.org} \\}

\begin{document}
\maketitle
\begin{abstract}
This paper evaluates the performance of transformer-based language models on split-ergative case alignment in Georgian, a particularly rare system for assigning grammatical cases to mark argument roles. We focus on \textit{subject} and \textit{object} marking determined through various permutations of \textit{nominative}, \textit{ergative}, and \textit{dative} noun forms. A treebank-based approach for the generation of minimal pairs using the Grew query language is implemented. We create a dataset of 370 syntactic tests made up of seven tasks containing 50--70 samples each, where three noun forms are tested in any given sample. Five encoder- and two decoder-only models are evaluated with word- and/or sentence-level accuracy metrics. Regardless of the specific syntactic makeup, models performed worst in assigning the ergative case correctly and strongest in assigning the nominative case correctly. Performance correlated with the overall frequency distribution of the three forms (\textsc{nom} $\textgreater$ \textsc{dat} $\textgreater$ \textsc{erg}). Though data scarcity is a known issue for low-resource languages, we show that the highly specific role of the ergative along with a lack of available training data likely contributes to poor performance on this case. The dataset is made publicly available and the methodology provides an interesting avenue for future syntactic evaluations of languages where benchmarks are limited.
\end{abstract}

\section{Introduction}
Small and Large Language models (LMs) have led to a paradigm shift in how we consider the learning of language by a machine. A natural question to ask has been: \textit{what understanding have they of the structure of the language they're trained on?} A standard approach to tackling this question has been the use of \textit{minimal pairs}, that is, grammatical/ungrammatical sentence pairs differing by a single syntactic feature. A significant effort has been put into compiling benchmarks for evaluating models on such pairs, however much of this has focused on a small subset of the world's approximately 7,000 languages. This has exacerbated a recognised dearth of research on \textit{low-resource languages} (LRLs), typically defined as languages that are under-studied, less commonly taught, or for which relatively few online resources exist \cite{less-privileged-langs, selection-lrls}. We consider Georgian one such language due to a significant lack of relevant research in this domain. This presents an opportunity however to analyse the rarer aspects of its syntax. 

In this work we evaluate one such aspect by looking at its split-ergative case-alignment system, exhibited in the same manner by only three other languages according to the \textit{World Atlas of Language Structures} (WALS; \citet{wals-98}). This revolves around three combinations of subject--object marking depending on verb tense/aspect/mood: nominative--dative, ergative--nominative, and dative--nominative. We use the Grew query language and Georgian Language Corpus Universal Dependencies treebank (GLC UD; \citet{georgian-treebank}) to generate minimal sets of syntactic tests. The resulting 370 case-alignment tests mark one of the first such datasets for Georgian\footnote{The full dataset is available on Hugging Face at DanielGallagherIRE/georgian-case-alignment.} and an evaluation of seven LMs reveals that models particularly struggle with the ergative case. Many LRLs suffer from a lack of available syntactic benchmarks and the approach used here may be generalised to other languages and phenomena where there is a treebank available.

\section{Related Work}
\label{sec:background-tse}
Transformer-based architectures \cite{attention} have become the standard for the training of high-performing language models and introduced a new wave of research in the field of Natural Language Processing (NLP). It appears that elements of syntax have been encoded by \textit{some} means, however our understanding of their sense of grammaticality is still limited \cite{kulmizev-schrodinger-2021}. There have been a variety of methods used to evaluate linguistic knowledge in LMs such as probing classifiers \citep{conneau-etal-2018-xnli, hewitt-manning-2019-structural, probing-classifiers-under-the-hood} or psycholinguistic tests \cite{psycholinguistic-tests}. Targeted Syntactic Evaluation (TSE) was introduced by \citet{marvin-linzen-2018-targeted} and brought a focus on minimal-pair acceptability judgements into the LM domain, where a minimal pair is defined as a set of two items with a single differing feature. In this context, the term is used to distinguish between a grammatical and ungrammatical sentence. Example \ref{background-tse-basic-example} shows agreement across a subject-relative clause with test \textit{run/runs}.
\pex
The officers that chased the thief \_.\\\\
p(\textit{run} | context) > p(\textit{runs} | context) \ding{51} \\
p(\textit{run} | context) $\leq$ p(\textit{runs} | context) \ding{55} \\
\label{background-tse-basic-example}
\xe
The \textit{Corpus of Linguistic Acceptability} (CoLA) was released soon after containing 10,657 sets of grammaticality judgements \citep{cola} and incorporated into the General Language Understanding Evaluation benchmark (GLUE; \citet{glue}). \citet{newman-2021-refining-tse} refined the goals of TSE further by focusing not just on how LMs fared with a single pair for a given sentence, but also expanded to include a wider variety of pairs beyond the most likely. It was observed that models could typically assign the correct conjugation to the expected verb, such as the distinction in Example \ref{background-tse-basic-example}, but struggled when it came to verbs that were less frequent despite the same grammatical/ungrammatical distinction e.g. \textit{rest/rests}. This provided evidence that models can have high syntactic performance on some tests without having generalised for that phenomenon across the language. Further work has been carried out on evaluating whether models have generalised syntactic phenomena across languages \cite{someya-etal-2024-targeted, hu-etal-2020-systematic}.

\subsection{Linguistic Minimal Pairs}
Minimal-pair syntactic tests have become the standard for evaluating model performance on syntactic phenomena and this ethos has resulted in a concerted effort towards the creation of standardised minimal-pair benchmarks. The \textit{Benchmark of Linguistic Minimal Pairs} (BLiMP; \citet{warstadt-blimp-2020}) compiled and published 67 datasets at 1000 sentences each for evaluating model performance on English syntax, remaining one of the most important datasets in the TSE domain. Similar benchmarks have been compiled for other languages such as German \citep{german-blimp}, Spanish \citep{spanish-blimp}, Chinese \citep{chinese-blimp, chinese-blimp-test}, Russian \citep{russian-blimp}, Japanese \citep{japanese-blimp}, and Dutch \citep{dutch-blimp}.

\subsection{Low-Resource Languages}
Work in the syntactic evaluation of LRLs has been limited. There has however been a notable buildup of momentum in the creation of BLiMP-style datasets, though with some being much smaller and covering fewer syntactic categories. These include datasets for Turkish \citep{turblimp}, Swedish \citep{swedish-blimp}, Icelandic \citep{icelandic-and-other-langs-blimp}, and more recently Irish \citep{irish-blimp}. \citet{evaluation-syntactic-knowledge-low-res} evaluated multilingual models on phenomena in three different languages: Swahili noun-class agreement, Hindi split ergativity, and Basque verb agreement. Performance correlated with how well a language was represented in the training data, with models performing strongest on Hindi. \citet{southeast-asian-minimal-pairs} focused on Southeast Asian languages and introduced a test suite for the evaluation of LMs on Tamil and Indonesian. A significant recent contribution is that of MultiBLiMP \citep{multi-blimp}, covering 101 languages with all data stemming from UD treebanks. They include Georgian as one of the languages including subject-verb and subject-participle agreement for both \texttt{number} and \texttt{person}, however they do not examine case alignment. Furthermore, they do not use a query language to match specific syntactic constructions but rather implement this programmatically. In this work we will take a query-based approach using Grew, a graph-based query language that can identify syntactic patterns in treebanks \cite{grew-1, grew-2}. 

\section{Georgian Syntax}
\label{sec:georgian}
Georgian is a Kartvelian language of the southern Caucasus syntactically characterised by its derivational morphology as well as its split-ergative case alignment system \cite{bolkvadze2023georgian}, the latter being discussed in detail in Section \ref{sec:split-ergative}. Modern Georgian uses the \textit{Mkhedruli} script written left-to-right and without distinction between upper and lower case \cite[p.~4]{book-georgian-grammar-ref}. An important aspect of this language's verb paradigm is that it does not distinguish clearly between \textit{tense}, \textit{aspect}, and \textit{mood}, instead categorising them into distinct properties known as \textit{screeves} \cite{georgian-thesis-reflex}. Henceforth we will refer to these tense--aspect--mood combinations using this term. 

The typical properties of an agglutinative language are exhibited with words built from an inventory of largely invariable morphemes \cite{agglutination-and-flection}. Unique suffixes represent any of seven noun cases as well as singular/plural marking that are appended to corresponding root forms \cite[p.~33]{book-georgian-grammar-ref}. A large variety of word forms are thus able to be produced. Only the nominative (\textsc{nom}), ergative (\textsc{erg}), and dative (\textsc{dat}) cases pertain to argument structure, therefore they will be our focus. How these cases are mapped to the argument roles determines the case-alignment system. We focus here solely on subject--object relations and will sometimes refer to particular alignments using \textsc{x--y} notation, where \textsc{x} represents the subject's case and \textsc{y} the object's.

\begin{table}[t!]
\centering
\begin{tabular}{llcc}
\toprule
Alignment & Screeve & Subject & Object \\ 
\midrule
Accusative  & \textsc{pres}/\textsc{fut} & \textsc{nom} & \textsc{dat} \\
Ergative  & \textsc{past} & \textsc{erg} & \textsc{nom} \\
Inverted  & \textsc{perf} & \textsc{dat} & \textsc{nom} \\
\bottomrule
\end{tabular}
\caption{Outline of the 3 Georgian alignments, their correspondings tense--aspect--mood category (i.e. screeve) triggers, and the resulting cases assigned to the subject and object in both transitive and intransitive constructions.}
\label{tbl:georgian-series-outline}
\end{table}

\subsection{Accusativity \& Ergativity}
\label{sec:ergativity}
The most common case-alignment system is \textit{accusative} \cite{wals-98}, where the nominative marks the subject of both transitive and intransitive verbs (e.g. \textit{he} runs, \textit{he} helps the boy) and the accusative marks the object of transitive verbs (e.g. he helps \textit{him}). Georgian however has no true accusative case and so the dative acts in its stead. In the screeves corresponding closely to the \textit{present} and \textit{future} tense we observe a nominative subject and dative object. Uniquely, in the \textit{perfective} screeve e.g. ``have done something'' (though requiring evidence of completion), these undergo inversion \cite[ch.~8]{harris1981georgian} and the subject is marked as dative and the object as nominative.
\begin{figure*}[t]
    \centering
    \includegraphics[width=1\linewidth]{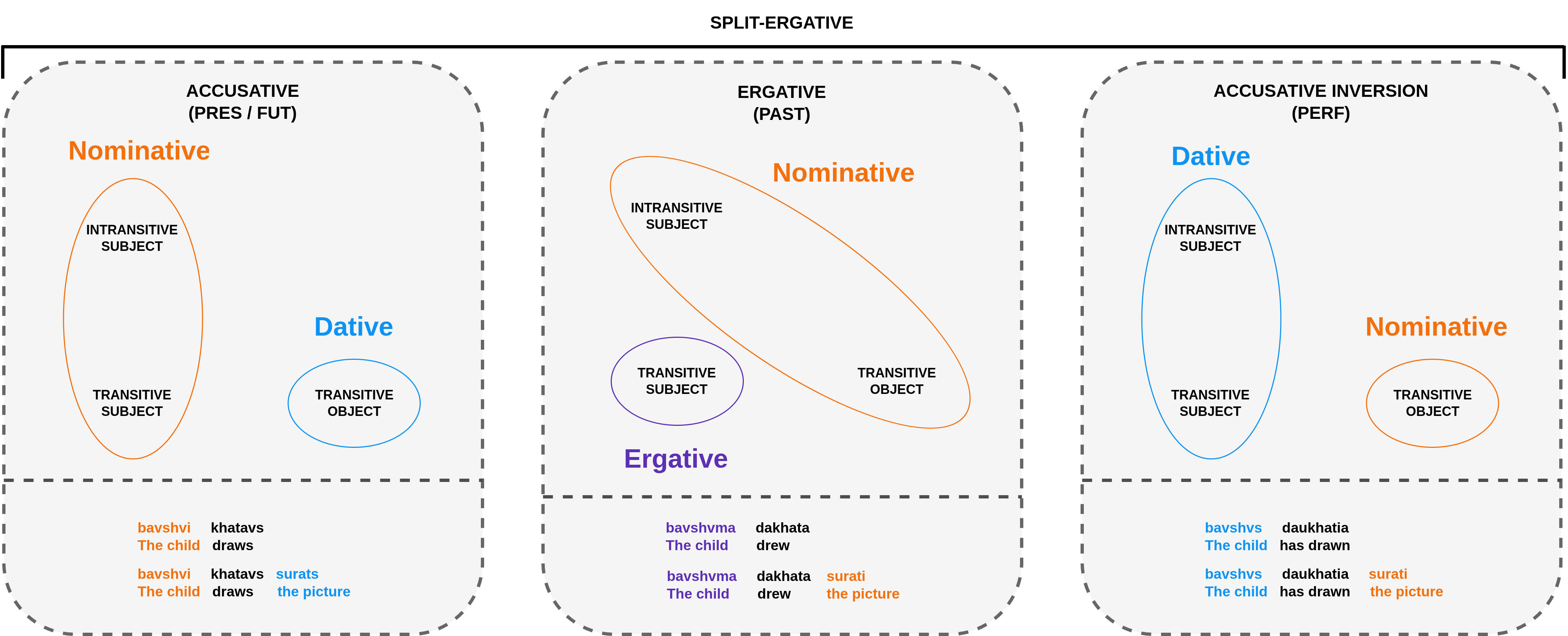}
    \caption{The Georgian split-ergative case system, made up of accusative, ergative, and inverted alignments. Examples are provided top-to-bottom for intransitive and transitive constructions. In most languages, including English, there is solely accusative alignment.}
    \label{fig:alignment-basic}
\end{figure*}
On the other hand, in the screeve corresponding to the \textit{past} tense, ergativity is observed. The key feature of this alignment is that a special case is assigned to the subject of transitive verbs. The subject of intransitive verbs and the object of transitive verbs thus take the same case, often referred to as the \textit{absolutive} case \cite{dixon1994ergativity} though we label it simply as the nominative. Table \ref{tbl:georgian-series-outline} shows an overview of the various alignments along with their corresponding screeves and assigned cases. 

\subsection{Split Ergativity}
\label{sec:split-ergative}
Ergative languages are often only partially ergative; under certain conditions they follow an ergative system and under others an accusative system \cite{split-ergativity-book}. If a language exhibits a mixture of alignments including at least partial ergativity, as is the case for Georgian, then this language typically falls under the category of \textit{split-ergative} \cite{interpretation-split-erg-delancey}. Screeves and their alignments are typically grouped into sets of `series' that are labelled I, II, and III \cite[ch.~1]{harris1981georgian}, however we will refer to the exact subject--object case alignments instead of individual series for simplicity. We leave the incorporation of further screeves as valuable future work. The full case-alignment system with corresponding examples is shown in Figure \ref{fig:alignment-basic}. The noun `bavshv' meaning `child' takes, for instance, three different cases when acting as the subject depending on the screeve: \textsc{nom}: `-i' , \textsc{erg}: `-ma' , or \textsc{dat}: `-s'. Example glosses are provided in Example \ref{gloss:georgian-basic}.

\pex\label{gloss:georgian-basic}
    \a
    \textbf{Present} \textsc{(nom)}
    \begingl
    \gla Bavshv\textbf{-i} tchams //
    \glb child-\textsc{nom} eat.\textsc{3sg.pres}//
    \glft ``The child eats.''//
    \endgl

    \a
    \textbf{Present} \textsc{(nom--dat)}
    \begingl
    \gla Mama sakhl\textbf{-s} ashenebs //
    \glb father.\textsc{nom} house-\textsc{dat} build.\textsc{3sg.pres}//
    \glft ``Father builds a house.''//
    \endgl

    \a
    \textbf{Past} \textsc{(erg--nom)}
    \begingl
    \gla Mama\textbf{-m} sakhl\textbf{-i} aashena  //
    \glb father-\textsc{erg} house-\textsc{nom} build.\textsc{3sg.pst}//
    \glft ``Father built a house.''//
    \endgl

    \a
    \textbf{Perfective} \textsc{(dat--nom)}
    \begingl
    \gla Mama\textbf{-s} sakhl\textbf{-i} aushenebia  //
    \glb father-\textsc{dat} house-\textsc{nom} build.\textsc{3sg.perf}//
    \glft ``Father has built a house (evidentially).''//
    \endgl
\xe
We can see from Figure \ref{fig:alignment-basic} that the same case may be used for a wide variety of syntactic or semantic roles and some naturally tend to occur more frequently than others. The Georgian nominative case, for instance, appears 11,438 times in the GLC UD treebank, while the dative 10,034 times and the ergative only 475 times. The nominative case thus occurs over 24 times as often as the ergative. This trend may persist in Georgian text generally.

\begin{table*}[t!]
\centering
\begin{tabular}{lcccc}
\toprule
Test Set & Subject & Object & Relation Tested & \# Tests \\ 
\midrule
\texttt{intransitive-nom-subj}  & Nominative & & Subject & 70 \\
\texttt{transitive-nom-dat-subj}  & Nominative & Dative & Subject & 50 \\
\texttt{transitive-nom-dat-obj}  & Nominative & Dative & Object & 50 \\
\texttt{transitive-erg-nom-subj}  & Ergative & Nominative & Subject & 50 \\
\texttt{transitive-erg-nom-obj}  & Ergative & Nominative & Object & 50 \\
\texttt{transitive-dat-nom-subj}  & Dative & Nominative & Subject & 50 \\
\texttt{transitive-dat-nom-obj}  & Dative & Nominative & Object & 50 \\
\bottomrule
\end{tabular}
\caption{Overview of the 7 Georgian case-alignment datasets from the GLC UD treebank totalling 370 tests. Table contains the dataset name, subject--object relations, relation tested, and number of tests.}
\label{tbl:dataset}
\end{table*}

\section{Data}
\label{sec:exp3-data}
We use Grew, a graph-based query language, for the creation of minimal-pair syntactic tests from the GLC UD treebank which contains 3,013 annotated sentences. Example \ref{app:grew-example-erg-nom} shows the queries used for a sample of the datasets. We search the treebank for four different forms of subject--object constructions: (1) intransitive nominative (2) transitive nominative--dative, (3) transitive ergative--nominative, and (4) transitive dative--nominative. These correspond respectively to accusative (for both (1) and (2)), ergative, and inverted alignments as discussed in Section \ref{sec:georgian}.

All data is in the Mkhedruli Georgian script as this is the expectation of model tokenisers. Table \ref{tbl:dataset} shows an outline of the datasets, with separate sets for testing the subject and object in transitive constructions resulting in a total of seven datasets of 50--70 samples each totalling 370 case-alignment tests. Each test sentence is accompanied with a target noun provided in the nominative, ergative, and dative form. Additional ergative forms were created and validated by a human annotator in order to reach at least 50 in any given set.
\begin{figure}[h!]
\centering
    \begin{lstlisting}[language=GREW, style=grewstyle]
            // Transitive Erg-Nom
            pattern {
              V [upos="VERB"];
              SUBJ [Case="Erg"];
              OBJ [Case="Nom"]; 
              V -[nsubj]-> SUBJ;
              V -[obj]-> OBJ;
            }

            // Intransitive Nom
            pattern {
                V [upos="VERB"];
                SUBJ [Case="Nom"];
                V -[nsubj]-> SUBJ;
            }        
            without {
                V [upos="VERB"];
                V -[nsubj]-> SUBJ;
                V -[obj]-> OBJ;
            }
    \end{lstlisting}
    \caption{Grew queries to match sentences in the GLC UD treebank that are transitive \textsc{erg-nom} and intransitive \textsc{nom} constructions.}
    \label{app:grew-example-erg-nom}
\end{figure}

\section{Methodology}
\label{sec:methodology}
We propose a query-based approach for generating syntactic tests from UD treebanks through the use of the Grew query language \footnote{Dataset creation and evaluation code is available on GitHub at DanielGall500/georgian-case-alignment.}. The usage of treebanks allows us to take advantage of (1) dependency relations between words for syntax specification, and (2) syntactic annotations. The latter allows us to find specific word forms and adjust them by a single morphosyntactic feature to create minimal pairs e.g. \textsc{nom} \textit{bavshvi} (lit. `child')$\rightarrow$\textsc{erg} ``\textit{bavshvma}''. Furthermore, the usage of a query language allows the intuitive specification of syntactic constructions. This approach relies heavily on the richness of the UD treebank in question, with some constructions and word forms naturally occurring less often than others, thus placing a limit on the number of syntactic tests that can be generated. In the case of Georgian, nominative and dative nouns occur frequently and ergative forms less so, therefore a native speaker provided assistance in creating additional ergative noun forms where necessary. Each test contains a masked sentence along with a nominative, dative, and ergative form of the same noun. Example tests with the three forms for transitive accusative and ergative alignments are shown in Example \ref{georgian:tse-examples}.
\pex\label{georgian:tse-examples}
    \a
    \begingl
    \gla bavshvi\text{\ding{51}}/bavshvma\text{\ding{55}}/bavshvs\text{\ding{55}} khatavs surats //
    \glb child.\textsc{nom}/child.\textsc{erg}/child.\textsc{dat} draws picture//
    \glft `The child draws a picture.'//
    \endgl

     \a
    \begingl
    \gla bavshvi\text{\ding{55}}/bavshvma\text{\ding{51}}/bavshvs\text{\ding{55}} dakhata  surati//
    \glb child.\textsc{nom}/child.\textsc{erg}/child.\textsc{dat} drew picture//
    \glft``The child drew a picture.''//
    \endgl
\label{dataset-samples}
\xe
\subsection{Minimal Sets}
The creation of a minimal pair using a treebank-generated lexicon requires (1) a base word form as it appears in text and (2) a single feature adjustment. In a typical TSE experiment configuration, two word forms are evaluated against each other, however we create a minimal set of three noun forms to account for all cases, one valid and two invalid in a given context. We thus make two feature adjustments, e.g. \textsc{ nom}$\rightarrow$\{\textsc{dat,erg}\}.

\subsection{Metrics}
\label{sec:metrics-accuracy}
The standard unit of measuring syntactic performance is \textit{accuracy}, defined as the proportion of answers that a model gets correct. Typically, accuracy relates to discrete or categorical values. The value with the highest probability is taken to be the final prediction $\hat{y}_i$ and compared with the ground truth $y_i$. We are not however comparing one discrete label with another but rather probabilities. For a given grammatical--ungrammatical pair ($w_G$, $w_{UG}$), a model is deemed correct if it assigns a higher probability to $w_G$ than to $w_{UG}$ given a context C \citep{marvin-linzen-2018-targeted}. Accuracy is defined in Equation \ref{eq:tse-accuracy} as applied across all tests. 
\begin{align}
\frac{1}{N} \sum_{i=1}^{N}
\mathbf{1}\Big[ P(w_G \mid C) > P(w_{UG} \mid C)\
\Big]
\label{eq:tse-accuracy}
\end{align}
Measuring accuracy in this manner may not account for the highest-probability token from a model inference as is otherwise typical. Rather, it compares any target words that have been pre-defined for the minimal pair. In our case we extrapolate this to the minimal \textit{set} of three forms where the grammatical item $w_G$ must be assigned a higher probability than the two ungrammatical items $w_{UG_{1}}$ and $w_{UG_{2}}$.

\subsubsection{Word- versus Sentence-level}
\label{sec:word-sent-level}
Comparing the likelihoods of words or sentences involves calculating the joint probabilities of their sequences of tokens. We implement two distinct approaches to the calculation of joint probabilities: \textit{sentence-level} (SL) and \textit{word-level} (WL). To calculate SL we take the joint probability over entire tokenised sentences auto-regressively. This can result in extremely low probabilities for long sentences, thus we convert this metric to a logarithmic scale to alleviate this issue. An additional problem with this metric is that the ungrammatical item may be present in the chain of probability multiplications long after the target word, causing unknown downstream effects on the results. WL evaluations, on the other hand, measure certainty at the exact moment that a grammatical/ungrammatical item is chosen by taking the joint probability of the tokenised target word only, given the full context.

\subsubsection{Why use one or the other?}
\label{why-wl-sl}
WL can be more representative of performance as it measures certainty at the exact moment that a model chooses an item in the minimal pair, whereas SL measures this across an entire tokenised sentence. However, in models trained on next-token prediction (i.e. decoder-only), WL only make sense if the grammaticality of a minimal pair is determined linearly \textit{earlier} in the sentence. Decoder-only models \textit{cannot} see the full context if the target is not the last word. For instance, in Georgian the verb often appears at the end of a sentence but the noun case is marked linearly earlier. SL metrics do not suffer from this as the full context is taken into account and thus ambiguities are limited. If we include decoder-only models, we should include SL evaluations to some extent despite WL being more representative of performance.

\begin{table*}[t!]
\centering
\makebox[\textwidth][c]{%
\begin{tabular}{llcccccccc}
\toprule
& & & \multicolumn{1}{c}{\textsc{nom}} & \multicolumn{2}{c}{\textsc{nom--dat}} & \multicolumn{2}{c}{\textsc{erg--nom}} & \multicolumn{2}{c}{\textsc{dat--nom}}\\
\cmidrule(lr){5-6}
\cmidrule(lr){7-8}
\cmidrule(lr){9-10}
Model 
& Size
& \#langs
& \textsc{subj} 
& \textsc{subj} 
& \textsc{obj} 
& \textsc{subj} 
& \textsc{obj} 
& \textsc{subj} 
& \textsc{obj} \\
\midrule
\multicolumn{10}{c}{Word-Level Accuracy (\%)} \\
\midrule

XLM-RoBERTa(bs) & 270M & 94 & \cellcolor{softgreen!73} 83 & \cellcolor{softgreen!86} 96 & \cellcolor{softgreen!42} 52 & \cellcolor{softgreen!16} 26 & \cellcolor{softgreen!74} 84 & \cellcolor{softgreen!46} 56 & \cellcolor{softgreen!86} 96 \\

XLM-RoBERTa(lg) & 550M & 94 & \cellcolor{softgreen!76} 86 & \cellcolor{softgreen!84} 94 & \cellcolor{softgreen!46} 56 & \cellcolor{softgreen!24} 34 & \cellcolor{softgreen!84} 94 & \cellcolor{softgreen!44} 54 & \cellcolor{softgreen!88} 98 \\

HPLT-BERT-ka & 110M & 1 & \cellcolor{softgreen!87} 97 & \cellcolor{softgreen!90} 100 & \cellcolor{softgreen!60} 70 & \cellcolor{softgreen!30} 40 & \cellcolor{softgreen!86} 96 & \cellcolor{softgreen!64} 74 & \cellcolor{softgreen!90} 100 \\

mBERT & 110M & 104 & \cellcolor{softgreen!90} 100 & \cellcolor{softgreen!82} 92 & \cellcolor{softgreen!32} 42 & \cellcolor{softgreen!0} 8 & \cellcolor{softgreen!88} 98 & \cellcolor{softgreen!26} 36 & \cellcolor{softgreen!88} 98 \\

RemBERT & 559M & 104 & \cellcolor{softgreen!49} 59 & \cellcolor{softgreen!62} 72 & \cellcolor{softgreen!10} 20 & \cellcolor{softgreen!4} 14 & \cellcolor{softgreen!62} 72 & \cellcolor{softgreen!18} 28 & \cellcolor{softgreen!46} 56 \\

\midrule

\multicolumn{10}{c}{Sentence-Level Accuracy (\%)} \\
\midrule

XLM-RoBERTa(bs) & 270M & 94 & \cellcolor{softgreen!40} 50 & \cellcolor{softgreen!44} 54 & \cellcolor{softgreen!8} 18 & \cellcolor{softgreen!4} 14 & \cellcolor{softgreen!42} 52 & \cellcolor{softgreen!14} 24 & \cellcolor{softgreen!52} 62 \\

XLM-RoBERTa(lg) & 550M & 94 & \cellcolor{softgreen!46} 56 & \cellcolor{softgreen!38} 48 & \cellcolor{softgreen!8} 18 & \cellcolor{softgreen!8} 18 & \cellcolor{softgreen!40} 50 & \cellcolor{softgreen!10} 20 & \cellcolor{softgreen!52} 62 \\

HPLT-BERT-ka & 110M & 1 & \cellcolor{softgreen!56} 66 & \cellcolor{softgreen!44} 54 & \cellcolor{softgreen!6} 16 & \cellcolor{softgreen!8} 18 & \cellcolor{softgreen!44} 54 & \cellcolor{softgreen!18} 28 & \cellcolor{softgreen!42} 52 \\

mBERT & 110M & 104 & \cellcolor{softgreen!61} 71 & \cellcolor{softgreen!58} 68 & \cellcolor{softgreen!0} 10 & \cellcolor{softgreen!4} 14 & \cellcolor{softgreen!74} 84 & \cellcolor{softgreen!4} 14 & \cellcolor{softgreen!70} 80 \\

RemBERT & 559M & 104 & \cellcolor{softgreen!51} 61 & \cellcolor{softgreen!52} 62 & \cellcolor{softgreen!20} 30 & \cellcolor{softgreen!16} 26 & \cellcolor{softgreen!60} 70 & \cellcolor{softgreen!14} 24 & \cellcolor{softgreen!50} 60 \\

GPT2-geo & 124M & 1 & \cellcolor{softgreen!59} 69 & \cellcolor{softgreen!58} 68 & \cellcolor{softgreen!28} 38 & \cellcolor{softgreen!6} 16 & \cellcolor{softgreen!60} 70 & \cellcolor{softgreen!24} 34 & \cellcolor{softgreen!56} 66 \\

mGPT-ka & 1.3B & 3 & \cellcolor{softgreen!87} 97 & \cellcolor{softgreen!90} 100 & \cellcolor{softgreen!90} 100 & \cellcolor{softgreen!74} 84 & \cellcolor{softgreen!88} 98 & \cellcolor{softgreen!68} 78 & \cellcolor{softgreen!90} 100 \\

\bottomrule

\end{tabular}
}
\caption{Accuracy scores for all models across tasks. \textsc{x--y} indicates the subject--object makeup and the tested role is underneath. Model performance is highest on the nominative regardless of the task and most models perform particularly poorly on assigning the ergative case.}
\label{tbl:results}
\end{table*}

\section{Experiments}
\label{sec:experiments}
We evaluate seven LMs, with encoder-only models evaluated on both WL and SL accuracy and decoder-only solely on SL accuracy for the reasons discussed in Section \ref{why-wl-sl}. The language marker for Georgian is `ka' or `geo'. The encoder-only models selected are XLM-RoBERTa base and large \citep{xlm-roberta}, BERT multilingual base \citep{mbert}, RemBERT \citep{rembert}, and HPLT-BERT-ka \citep{hplt-1, hplt-2}. All are multilingual except for the HPLT model which is trained solely on Georgian. The decoder-only models are GPT2-geo \citep{gpt2-geo} and mGPT-ka \citep{mgpt, mgpt-geo}, both models fine-tuned on Georgian though with the latter additionally fine-tuned on English and Russian. Models range from 110M--1.3B parameters and 1--104 languages. They use a variety of tokenisation algorithms consisting of WordPiece \cite{wordpiece} for mBERT and HPLT-BERT-ka, SentencePiece \cite{sentencepiece} for RemBERT and the XLM-R models, and Byte-Pair Encoding (BPE; \citet{bpe}) for the GPT models.

\section{Results}
\label{sec:results}
\begin{figure}
    \centering
    \includegraphics[width=0.9\linewidth]{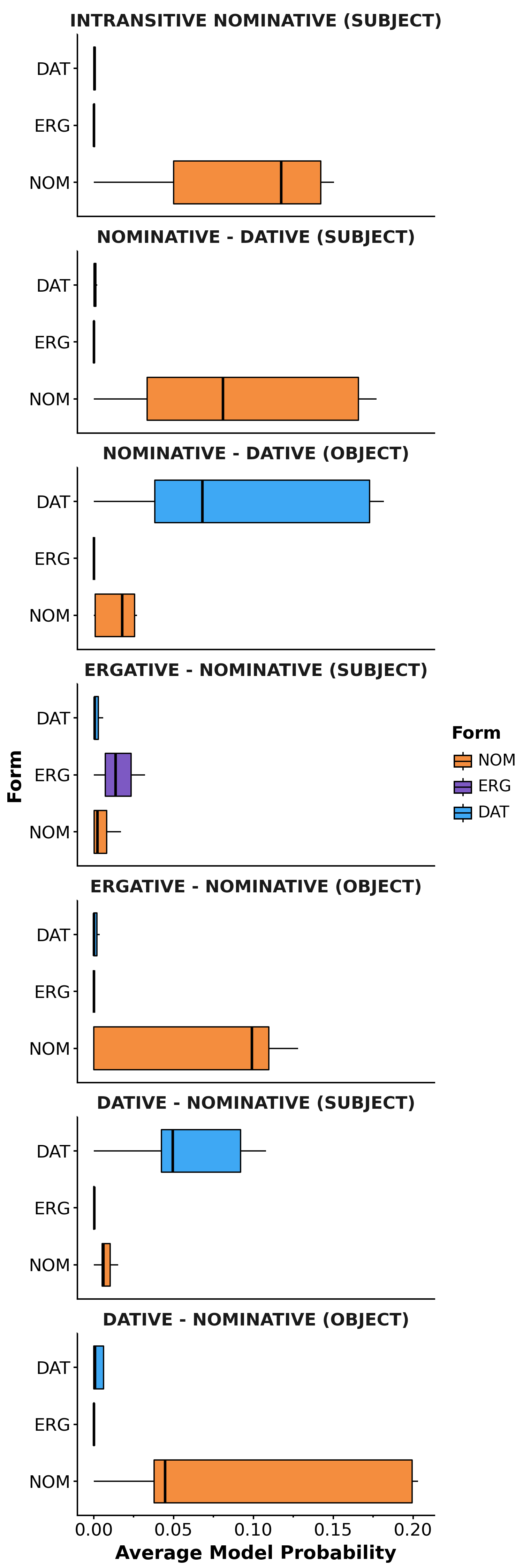}
    \caption{The average word-level probability assigned to each case represented as box plots. In all cases, the highest average probability is assigned to the correct grammatical case. Most significantly however, the average probability is significantly lower where the model must assign the ergative case correctly.}
    \label{fig:plot-avg-prob}
\end{figure}

The resulting WL and SL accuracy scores are shown in Table \ref{tbl:results}. The cases assigned are indicated in the \textsc{x--y} format and the \textsc{subj} or \textsc{obj} label indicates which role was tested. Figure \ref{fig:plot-avg-prob} shows box plots with the average probability assigned to each case for each dataset, listed in the same order top-to-bottom as in Table \ref{tbl:results}. Note that this shows the results for the word-level only as they are more comparable due to fewer probability multiplications. The results show the strongest overall performance on assigning the nominative correctly, followed by notably poorer performance on dative and ergative. Across all tasks that tested the correct assignment of the nominative, the average word- and sentence-level accuracy was 88.6\% and 67.3\%, respectively. This was followed by 48.8\% and 32.3\% for the dative, and 24.4\% and 27.1\% for the ergative. The majority of models perform poorest on the ergative. Similarly, Figure \ref{fig:plot-avg-prob} shows a dramatic reduction in model certainty for WL evaluations when evaluating the grammatical ergative form. On the other hand, models assign a much higher average probability to the correct form for tasks that test the nominative and dative. This effect persists regardless of the syntactic task.

\subsection{Discussion}
The strongest performance across tasks is observed from the mGPT-ka decoder-only model, notably the largest tested. High model accuracy alone however tells us little of the generalisation of a syntactic feature due to issues with testing on unseen data. Poor performance, on the other hand, \textit{does} indicate a meaningful gap in knowledge. We thus see the almost universally poor performance assigning the ergative, discussed further in Section \ref{sec:ergative-lacking}, as of particular interest in these results. We additionally observe a notable nominative bias that extends across \textit{all} tasks, indicating an effect that is beyond simple memorisation. We will explore this in Section \ref{sec:nominative-default-bias}. 

The frequency distribution of Georgian case approximately follows \textsc{nom} $\textgreater$ \textsc{dat} $\textgreater$ \textsc{erg} (see Section \ref{sec:georgian}). There is an interesting correlation between this general frequency of cases and the word-level accuracy, sentence-level accuracy, and probability results: \textsc{nom} (\textsc{wl}: 88.6\%, \textsc{sl}: 67.3\%, \textsc{wl p(x)}: 0.03) $\textgreater$ \textsc{dat} (\textsc{wl}: 32.3\%, \textsc{sl}: 36.6\%, \textsc{wl p(x)}: 0.029) $\textgreater$ \textsc{erg} (\textsc{wl}: 24.4\%, \textsc{sl}: 27.1\%, \textsc{wl p(x)}: 0.008). These results indicate a possible relationship between the general frequency of syntactic constructions and their learnability in data-scarce contexts.

TSE results can be difficult to interpret (see Section \ref{sec:background-tse}) and we have included multiple accuracy metrics to alleviate this. We will however focus primarily on word-level metrics to identify patterns in the results due to their more direct measurement of model preference (see Section \ref{sec:word-sent-level}). We will additionally assess the possible impact of tokenised word or sentence length on performance in Section \ref{sec:tokenisation-assessment}.

\subsection{Models Struggle with Ergativity}
\label{sec:ergative-lacking}
A key finding is that models tend to struggle most with correctly assigning the ergative case. The nominative and dative are used for a wider variety of purposes and naturally occur more frequently, while the ergative is used in very specific cases (sometimes defined as a more \textit{marked} feature \cite{markedness}). There is thus a natural reduction in instances of the ergative case in the training data, strained further by limited data for this language generally. This leads us to a situation where models may easily over-generalise and thus fail to learn the correct triggers for the ergative case, namely, the usage of the perfective screeve. Furthermore, the rarity of this case-alignment system across languages likely means that there is limited scope for improvement through transfer learning. Data scarcity is a well-understood issue for training models on LRLs \cite{multilinguality-curse, overcomingdatascarcity}, however here we have isolated its effects on the learning of syntax to a single syntactic feature.

\subsection{Nominative Bias}
\label{sec:nominative-default-bias}
There is conversely a clear default bias towards the nominative case regardless of syntactic makeup. In cases where the dative was the correct choice but models chose wrongly, the nominative was preferred 83.2\% of the time with only minor deviation among datasets ($\sigma$=2\%). In failed tests for the ergative case, the nominative was preferred an average of 74.4\% of the time. Lastly, where the nominative was the correct choice but the model chose wrongly, the dative was preferred over the ergative an average of 79\% of the time with limited deviation across the four datasets ($\sigma$=3\%). Even where models did not choose the nominative as they should have, the dative is a secondary default and the ergative is rarely preferred.

\subsection{Assessing Tokenisation Impact}
\label{sec:tokenisation-assessment}
Agglutinative languages such as Turkish or Georgian provide particular challenges for tokenisation algorithms due to the building of particularly complex word forms \cite{tokeneffect-lang}. However, we found that their effect on the end result was in this case limited and, furthermore, we did not find a qualitative word or sentence type that correlated with poor performance. Models used a variety of tokenisation approaches (see Section \ref{sec:experiments}) and the average token length of the target word for the WL results differed significantly: mBERT -- 4.1 $\textgreater$ RemBERT -- 2.3 $\textgreater$ XLM-R(lg) -- 2.2 $\textgreater$ XLM-R(bs) -- 2.1, $\textgreater$ HPLT-BERT-ka -- 1.4. Upon performing a Pearson correlation analysis between model performance and both tokenised (a) target word length, and (b) sentence length, we found no statistically significant correlations for the nominative (a. $r$=--0.34, $p$=0.21; b. $r$=0.04, $p$=0.45), the dative (a. $r$=--0.24, $p$=0.2; b. $r$=--0.04, $p$=0.53), or the ergative (a. $r$=0.03, p=0.34; b. $r$=0.07, $p$=0.42) tasks. These findings partially alleviate the limitation that we did not control for sentence length and point further toward the rarity of the ergative and general data scarcity for Georgian as a key issue.

\section{Conclusion}
In this work we created a dataset for the syntactic evaluation of Georgian case alignment and evaluated seven LMs trained on Georgian text. Performance correlated with the distribution of case seen within the language: \textsc{nom} $\textgreater$ \textsc{dat} $\textgreater$ \textsc{erg}. The low frequency of the ergative case, its highly specific use case, as well as an overall lack of available training data make this a particularly difficult syntactic phenomenon for models to learn. We found that models tended to default to using the nominative regardless of the syntactic construction as shown by an error analysis of model preferences. Furthermore, a Pearson correlation analysis revealed that tokenised word and sentence length did not appear to be strong indicators of performance. The dataset, primarily generated from the GLC UD treebank, contributes towards a syntactic benchmark for Georgian. The approach used may also be generalised to other languages for which a treebank exists. The incorporation into the dataset of additional screeves as well as a broader range of intransitive tests would be a fruitful avenue for future work.

\section*{Limitations}
A treebank-based approach to the creation of minimal pairs comes with the caveat that any generated tests have possibly been seen in the training data, leading to difficulties in properly evaluating the generalisation of syntactic features or performance on unseen data. Furthermore, an upper limit is placed on the number of syntactic tests that can be generated for a given feature that is determined by the treebank size. Dataset size was thus a limiting factor in this work. Due to these limitations we additionally did not control for sentence length, though this was shown not to have major implications in Section \ref{sec:tokenisation-assessment}. There are additional difficulties in the interpretation of the results due to tokenisation biases and context ambiguities.

\section*{Acknowledgments}
A special thank you to Dr. Tamara Tartarashvili for her validation of the Georgian dataset as well as Christopher Schröder, Samantha Zielinski, and the anonymous reviewers for their helpful feedback. Part of this work was conducted within the \href{https://coral-nlp.github.io}{CORAL project} funded by the German Federal Ministry of Research, Technology, and Space (BMFTR) under the grant number 16IS24077A. Responsibility for the content of this publication lies with the authors.

\bibliography{custom}

@inproceedings{georgian-treebank,
    title = "Building a {U}niversal {D}ependencies Treebank for {G}eorgian",
    author = "Lobzhanidze, Irina  and
      Magradze, Erekle  and
      Berikashvili, Svetlana  and
      Gozalishvili, Anzor  and
      Jalaghonia, Tamar",
    editor = {Dakota, Daniel  and
      Jablotschkin, Sarah  and
      K{\"u}bler, Sandra  and
      Zinsmeister, Heike},
    booktitle = "Proceedings of the 22nd Workshop on Treebanks and Linguistic Theories (TLT 2024)",
    month = dec,
    year = "2024",
    address = "Hamburg,Germany",
    publisher = "Association for Computational Linguistics",
    url = "https://aclanthology.org/2024.tlt-1.5/",
    pages = "40--45"
}

@book{dixon1994ergativity,
  title={Ergativity},
  author={Dixon, R.M.W.},
  isbn={9780521448987},
  lccn={93015925},
  series={Cambridge Studies in Linguistics},
  url={https://books.google.de/books?id=fKfSAu6v5LYC},
  year={1994},
  publisher={Cambridge University Press}
}

@incollection{agglutination-and-flection,
  author    = {Plungian, Vladimir},
  title     = {Agglutination and Flection},
  booktitle = {Language Typology and Language Universals: An International Handbook},
  volume    = {1},
  pages     = {669--678},
  editor    = {Haspelmath, Martin and others},
  year      = {2001},
  publisher = {Mouton de Gruyter},
  address   = {Berlin}
}

@article{book-georgian-grammar-ref,
author = {Tuite, Kevin},
year = {1996},
month = {01},
pages = {},
title = {B. G. Hewitt. Georgian: A Structural Reference Grammar},
volume = {3},
journal = {Functions of Language},
doi = {10.1075/fol.3.2.07tui}
}

@book{harris1981georgian,
  title={Georgian Syntax: A Study in Relational Grammar},
  author={Harris, A.C.},
  isbn={9780521235846},
  lccn={80041497},
  series={Cambridge Studies in Linguistics},
  url={https://books.google.de/books?id=qxQsWirVoSwC},
  year={1981},
  publisher={Cambridge University Press}
}

@INPROCEEDINGS{wordpiece,

  author={Schuster, Mike and Nakajima, Kaisuke},

  booktitle={2012 IEEE International Conference on Acoustics, Speech and Signal Processing (ICASSP)}, 

  title={Japanese and Korean voice search}, 

  year={2012},

  volume={},

  number={},

  pages={5149-5152},

  doi={10.1109/ICASSP.2012.6289079}}

@inproceedings{bpe,
    title = "Neural Machine Translation of Rare Words with Subword Units",
    author = "Sennrich, Rico  and
      Haddow, Barry  and
      Birch, Alexandra",
    editor = "Erk, Katrin  and
      Smith, Noah A.",
    booktitle = "Proceedings of the 54th Annual Meeting of the Association for Computational Linguistics (Volume 1: Long Papers)",
    month = aug,
    year = "2016",
    address = "Berlin, Germany",
    publisher = "Association for Computational Linguistics",
    url = "https://aclanthology.org/P16-1162/",
    doi = "10.18653/v1/P16-1162",
    pages = "1715--1725"
}

@inproceedings{sentencepiece,
    title = "{S}entence{P}iece: A simple and language independent subword tokenizer and detokenizer for Neural Text Processing",
    author = "Kudo, Taku  and
      Richardson, John",
    editor = "Blanco, Eduardo  and
      Lu, Wei",
    booktitle = "Proceedings of the 2018 Conference on Empirical Methods in Natural Language Processing: System Demonstrations",
    month = nov,
    year = "2018",
    address = "Brussels, Belgium",
    publisher = "Association for Computational Linguistics",
    url = "https://aclanthology.org/D18-2012/",
    doi = "10.18653/v1/D18-2012",
    pages = "66--71"
}

@misc{tokeneffect-lang,
      title={TokSuite: Measuring the Impact of Tokenizer Choice on Language Model Behavior}, 
      author={Gül Sena Altıntaş and Malikeh Ehghaghi and Brian Lester and Fengyuan Liu and Wanru Zhao and Marco Ciccone and Colin Raffel},
      year={2025},
      eprint={2512.20757},
      archivePrefix={arXiv},
      primaryClass={cs.CL},
      url={https://arxiv.org/abs/2512.20757}, 
}

@inproceedings{marvin-linzen-2018-targeted,
    title = "Targeted Syntactic Evaluation of Language Models",
    author = "Marvin, Rebecca  and
      Linzen, Tal",
    editor = "Riloff, Ellen  and
      Chiang, David  and
      Hockenmaier, Julia  and
      Tsujii, Jun{'}ichi",
    booktitle = "Proceedings of the 2018 Conference on Empirical Methods in Natural Language Processing",
    month = oct # "-" # nov,
    year = "2018",
    address = "Brussels, Belgium",
    publisher = "Association for Computational Linguistics",
    url = "https://aclanthology.org/D18-1151/",
    doi = "10.18653/v1/D18-1151",
    pages = "1192--1202"
}

@inproceedings{glue,
    title = "{GLUE}: A Multi-Task Benchmark and Analysis Platform for Natural Language Understanding",
    author = "Wang, Alex  and
      Singh, Amanpreet  and
      Michael, Julian  and
      Hill, Felix  and
      Levy, Omer  and
      Bowman, Samuel",
    editor = "Linzen, Tal  and
      Chrupa{\l}a, Grzegorz  and
      Alishahi, Afra",
    booktitle = "Proceedings of the 2018 {EMNLP} Workshop {B}lackbox{NLP}: Analyzing and Interpreting Neural Networks for {NLP}",
    month = nov,
    year = "2018",
    address = "Brussels, Belgium",
    publisher = "Association for Computational Linguistics",
    url = "https://aclanthology.org/W18-5446/",
    doi = "10.18653/v1/W18-5446",
    pages = "353--355"
}

@misc{gpt2-geo,
  author       = {Kuduxaaa},
  title        = {{GPT-2Geo: Georgian Language Model}},
  year         = {2025},
  howpublished = {\url{https://huggingface.co/Kuduxaaa/gpt2-geo}},
  note         = {Accessed: 2025-12-26}
}

@misc{mgpt-geo,
  author       = {AI-Forever},
  title        = {{mGPT-1.3B-Georgian: Georgian Language Model}},
  year         = {2025},
  howpublished = {\url{https://huggingface.co/ai-forever/mGPT-1.3B-georgian}},
  note         = {Accessed: 2025-12-26}
}

@misc{southeast-asian-minimal-pairs,
      title={BHASA: A Holistic Southeast Asian Linguistic and Cultural Evaluation Suite for Large Language Models}, 
      author={Wei Qi Leong and Jian Gang Ngui and Yosephine Susanto and Hamsawardhini Rengarajan and Kengatharaiyer Sarveswaran and William Chandra Tjhi},
      year={2023},
      eprint={2309.06085},
      archivePrefix={arXiv},
      primaryClass={cs.CL},
      url={https://arxiv.org/abs/2309.06085}, 
}

@misc{warstadt-blimp-2020,
    title = "{BL}i{MP}: The Benchmark of Linguistic Minimal Pairs for {E}nglish",
    author = "Warstadt, Alex  and
      Parrish, Alicia  and
      Liu, Haokun  and
      Mohananey, Anhad  and
      Peng, Wei  and
      Wang, Sheng-Fu  and
      Bowman, Samuel R.",
    editor = "Johnson, Mark  and
      Roark, Brian  and
      Nenkova, Ani",
    journal = "Transactions of the Association for Computational Linguistics",
    volume = "8",
    year = "2020",
    address = "Cambridge, MA",
    publisher = "MIT Press",
    url = "https://aclanthology.org/2020.tacl-1.25/",
    doi = "10.1162/tacl_a_00321",
    pages = "377--392"
}

@inproceedings{attention,
author = {Vaswani, Ashish and Shazeer, Noam and Parmar, Niki and Uszkoreit, Jakob and Jones, Llion and Gomez, Aidan N. and Kaiser, \L{}ukasz and Polosukhin, Illia},
title = {Attention is all you need},
year = {2017},
isbn = {9781510860964},
publisher = {Curran Associates Inc.},
address = {Red Hook, NY, USA},
booktitle = {Proceedings of the 31st International Conference on Neural Information Processing Systems},
pages = {6000–6010},
numpages = {11},
location = {Long Beach, California, USA},
series = {NIPS'17}
}

@inproceedings{grew-1,
  TITLE = {{Graph Matching and Graph Rewriting: GREW tools for corpus exploration, maintenance and conversion}},
  AUTHOR = {Guillaume, Bruno},
  URL = {https://inria.hal.science/hal-03177701},
  BOOKTITLE = {{EACL 2021 - 16th conference of the European Chapter of the Association for Computational Linguistics: System Demonstrations}},
  ADDRESS = {Kiev/Online, Ukraine},
  YEAR = {2021},
  MONTH = Apr,
  HAL_ID = {hal-03177701},
  HAL_VERSION = {v1},
}

@inproceedings{grew-2,
  TITLE = {{Graph Matching for Corpora Exploration}},
  AUTHOR = {Guillaume, Bruno},
  URL = {https://inria.hal.science/hal-02267475},
  BOOKTITLE = {{JLC 2019 - 10{\`e}mes Journ{\'e}es Internationales de la Linguistique de corpus}},
  ADDRESS = {Grenoble, France},
  YEAR = {2019},
  MONTH = Nov,
  HAL_ID = {hal-02267475},
  HAL_VERSION = {v1},
}

@phdthesis{georgian-thesis-reflex,
  author       = {Nino Amiridze},
  title        = {Reflexivization Strategies in Georgian},
  school       = {Utrecht University},
  year         = {2006},
  series       = {LOT Dissertation Series},
  volume       = {127},
  address      = {Utrecht, The Netherlands},
  publisher    = {LOT Publications},
}

@book{markedness,
  title={Markedness theory},
  author={Andrews, Edna},
  year={1990},
  publisher={Duke University Press}
}

@article{screeves,
  title={Section 5. Languages of the world},
  author={Jikurashvili, Tinatin}
}

@inproceedings{hplt-1,
    title = "Trained on 100 million words and still in shape: {BERT} meets {B}ritish {N}ational {C}orpus",
    author = "Samuel, David  and
      Kutuzov, Andrey  and
      {\O}vrelid, Lilja  and
      Velldal, Erik",
    editor = "Vlachos, Andreas  and
      Augenstein, Isabelle",
    booktitle = "Findings of the Association for Computational Linguistics: EACL 2023",
    month = may,
    year = "2023",
    address = "Dubrovnik, Croatia",
    publisher = "Association for Computational Linguistics",
    url = "https://aclanthology.org/2023.findings-eacl.146",
    doi = "10.18653/v1/2023.findings-eacl.146",
    pages = "1954--1974"
}

@inproceedings{hplt-2,
    title = "A New Massive Multilingual Dataset for High-Performance Language Technologies",
    author = {de Gibert, Ona  and
      Nail, Graeme  and
      Arefyev, Nikolay  and
      Ba{\~n}{\'o}n, Marta  and
      van der Linde, Jelmer  and
      Ji, Shaoxiong  and
      Zaragoza-Bernabeu, Jaume  and
      Aulamo, Mikko  and
      Ram{\'\i}rez-S{\'a}nchez, Gema  and
      Kutuzov, Andrey  and
      Pyysalo, Sampo  and
      Oepen, Stephan  and
      Tiedemann, J{\"o}rg},
    editor = "Calzolari, Nicoletta  and
      Kan, Min-Yen  and
      Hoste, Veronique  and
      Lenci, Alessandro  and
      Sakti, Sakriani  and
      Xue, Nianwen",
    booktitle = "Proceedings of the 2024 Joint International Conference on Computational Linguistics, Language Resources and Evaluation (LREC-COLING 2024)",
    month = may,
    year = "2024",
    address = "Torino, Italia",
    publisher = "ELRA and ICCL",
    url = "https://aclanthology.org/2024.lrec-main.100",
    pages = "1116--1128",
}

@inproceedings{hu-etal-2020-systematic,
    title = "A Systematic Assessment of Syntactic Generalization in Neural Language Models",
    author = "Hu, Jennifer  and
      Gauthier, Jon  and
      Qian, Peng  and
      Wilcox, Ethan  and
      Levy, Roger",
    editor = "Jurafsky, Dan  and
      Chai, Joyce  and
      Schluter, Natalie  and
      Tetreault, Joel",
    booktitle = "Proceedings of the 58th Annual Meeting of the Association for Computational Linguistics",
    month = jul,
    year = "2020",
    address = "Online",
    publisher = "Association for Computational Linguistics",
    url = "https://aclanthology.org/2020.acl-main.158/",
    doi = "10.18653/v1/2020.acl-main.158",
    pages = "1725--1744"
}

@misc{overcomingdatascarcity,
      title={Overcoming Data Scarcity in Generative Language Modelling for Low-Resource Languages: A Systematic Review}, 
      author={Josh McGiff and Nikola S. Nikolov},
      year={2025},
      eprint={2505.04531},
      archivePrefix={arXiv},
      primaryClass={cs.CL},
      url={https://arxiv.org/abs/2505.04531}, 
}

@inproceedings{multilinguality-curse,
    title = "When Is Multilinguality a Curse? Language Modeling for 250 High- and Low-Resource Languages",
    author = "Chang, Tyler A.  and
      Arnett, Catherine  and
      Tu, Zhuowen  and
      Bergen, Benjamin K.",
    editor = "Al-Onaizan, Yaser  and
      Bansal, Mohit  and
      Chen, Yun-Nung",
    booktitle = "Proceedings of the 2024 Conference on Empirical Methods in Natural Language Processing",
    month = nov,
    year = "2024",
    address = "Miami, Florida, USA",
    publisher = "Association for Computational Linguistics",
    url = "https://aclanthology.org/2024.emnlp-main.236/",
    doi = "10.18653/v1/2024.emnlp-main.236",
    pages = "4074--4096"
}

@book{split-ergativity-book,
  title={Aspects of split ergativity},
  author={Coon, Jessica},
  year={2013},
  publisher={Oxford University Press}
}

@book{bolkvadze2023georgian,
  title={Georgian: A comprehensive grammar},
  author={Bolkvadze, Tinatin and Kiziria, Dodona},
  year={2023},
  publisher={Routledge}
}

@article{interpretation-split-erg-delancey,
 ISSN = {00978507, 15350665},
 URL = {http://www.jstor.org/stable/414343},
 author = {Scott DeLancey},
 journal = {Language},
 number = {3},
 pages = {626--657},
 publisher = {Linguistic Society of America},
 title = {An Interpretation of Split Ergativity and Related Patterns},
 urldate = {2026-01-06},
 volume = {57},
 year = {1981}
}

@incollection{wals-98,
  author    = {Bernard Comrie},
  booktitle = {The World Atlas of Language Structures Online},
  editor    = {Matthew S. Dryer and Martin Haspelmath},
  publisher = {Zenodo},
  title     = {Alignment of Case Marking of Full Noun Phrases (v2020.4)},
  type      = {Data set},
  url       = {https://doi.org/10.5281/zenodo.13950591},
  year      = {2013},
  doi       = {10.5281/zenodo.13950591}
}

@inproceedings{less-privileged-langs,
    title = "Natural Language Processing for Less Privileged Languages: Where do we come from? Where are we going?",
    author = "Singh, Anil Kumar",
    booktitle = "Proceedings of the {IJCNLP}-08 Workshop on {NLP} for Less Privileged Languages",
    year = "2008",
    url = "https://aclanthology.org/I08-3004/"
}

@inproceedings{selection-lrls,
    title = "Selection Criteria for Low Resource Language Programs",
    author = "Cieri, Christopher  and
      Maxwell, Mike  and
      Strassel, Stephanie  and
      Tracey, Jennifer",
    editor = "Calzolari, Nicoletta  and
      Choukri, Khalid  and
      Declerck, Thierry  and
      Goggi, Sara  and
      Grobelnik, Marko  and
      Maegaard, Bente  and
      Mariani, Joseph  and
      Mazo, Helene  and
      Moreno, Asuncion  and
      Odijk, Jan  and
      Piperidis, Stelios",
    booktitle = "Proceedings of the Tenth International Conference on Language Resources and Evaluation ({LREC}'16)",
    month = may,
    year = "2016",
    address = "Portoro{\v{z}}, Slovenia",
    publisher = "European Language Resources Association (ELRA)",
    url = "https://aclanthology.org/L16-1720/",
    pages = "4543--4549"
}

@inproceedings{conneau-etal-2018-xnli,
    title = "{XNLI}: Evaluating Cross-lingual Sentence Representations",
    author = "Conneau, Alexis  and
      Rinott, Ruty  and
      Lample, Guillaume  and
      Williams, Adina  and
      Bowman, Samuel  and
      Schwenk, Holger  and
      Stoyanov, Veselin",
    editor = "Riloff, Ellen  and
      Chiang, David  and
      Hockenmaier, Julia  and
      Tsujii, Jun{'}ichi",
    booktitle = "Proceedings of the 2018 Conference on Empirical Methods in Natural Language Processing",
    month = oct # "-" # nov,
    year = "2018",
    address = "Brussels, Belgium",
    publisher = "Association for Computational Linguistics",
    url = "https://aclanthology.org/D18-1269/",
    doi = "10.18653/v1/D18-1269",
    pages = "2475--2485"
}

@inproceedings{hewitt-manning-2019-structural,
    title = "{A} Structural Probe for Finding Syntax in Word Representations",
    author = "Hewitt, John  and
      Manning, Christopher D.",
    editor = "Burstein, Jill  and
      Doran, Christy  and
      Solorio, Thamar",
    booktitle = "Proceedings of the 2019 Conference of the North {A}merican Chapter of the Association for Computational Linguistics: Human Language Technologies, Volume 1 (Long and Short Papers)",
    month = jun,
    year = "2019",
    address = "Minneapolis, Minnesota",
    publisher = "Association for Computational Linguistics",
    url = "https://aclanthology.org/N19-1419/",
    doi = "10.18653/v1/N19-1419",
    pages = "4129--4138"
}

@inproceedings{psycholinguistic-tests,
    title = "Neural language models as psycholinguistic subjects: Representations of syntactic state",
    author = "Futrell, Richard  and
      Wilcox, Ethan  and
      Morita, Takashi  and
      Qian, Peng  and
      Ballesteros, Miguel  and
      Levy, Roger",
    editor = "Burstein, Jill  and
      Doran, Christy  and
      Solorio, Thamar",
    booktitle = "Proceedings of the 2019 Conference of the North {A}merican Chapter of the Association for Computational Linguistics: Human Language Technologies, Volume 1 (Long and Short Papers)",
    month = jun,
    year = "2019",
    address = "Minneapolis, Minnesota",
    publisher = "Association for Computational Linguistics",
    url = "https://aclanthology.org/N19-1004/",
    doi = "10.18653/v1/N19-1004",
    pages = "32--42"
}

@article{mgpt,
    title = "m{GPT}: Few-Shot Learners Go Multilingual",
    author = "Shliazhko, Oleh  and
      Fenogenova, Alena  and
      Tikhonova, Maria  and
      Kozlova, Anastasia  and
      Mikhailov, Vladislav  and
      Shavrina, Tatiana",
    journal = "Transactions of the Association for Computational Linguistics",
    volume = "12",
    year = "2024",
    address = "Cambridge, MA",
    publisher = "MIT Press",
    url = "https://aclanthology.org/2024.tacl-1.4/",
    doi = "10.1162/tacl_a_00633",
    pages = "58--79"
}

@misc{rembert,
      title={Rethinking embedding coupling in pre-trained language models}, 
      author={Hyung Won Chung and Thibault Févry and Henry Tsai and Melvin Johnson and Sebastian Ruder},
      year={2020},
      eprint={2010.12821},
      archivePrefix={arXiv},
      primaryClass={cs.CL},
      url={https://arxiv.org/abs/2010.12821}, 
}

@article{mbert,
  author    = {Jacob Devlin and
               Ming{-}Wei Chang and
               Kenton Lee and
               Kristina Toutanova},
  title     = {{BERT:} Pre-training of Deep Bidirectional Transformers for Language
               Understanding},
  journal   = {CoRR},
  volume    = {abs/1810.04805},
  year      = {2018},
  url       = {http://arxiv.org/abs/1810.04805},
  archivePrefix = {arXiv},
  eprint    = {1810.04805},
  bibsource = {dblp computer science bibliography, https://dblp.org}
}

@inproceedings{xlm-roberta,
    title = "Unsupervised Cross-lingual Representation Learning at Scale",
    author = "Conneau, Alexis  and
      Khandelwal, Kartikay  and
      Goyal, Naman  and
      Chaudhary, Vishrav  and
      Wenzek, Guillaume  and
      Guzm{\'a}n, Francisco  and
      Grave, Edouard  and
      Ott, Myle  and
      Zettlemoyer, Luke  and
      Stoyanov, Veselin",
    editor = "Jurafsky, Dan  and
      Chai, Joyce  and
      Schluter, Natalie  and
      Tetreault, Joel",
    booktitle = "Proceedings of the 58th Annual Meeting of the Association for Computational Linguistics",
    month = jul,
    year = "2020",
    address = "Online",
    publisher = "Association for Computational Linguistics",
    url = "https://aclanthology.org/2020.acl-main.747/",
    doi = "10.18653/v1/2020.acl-main.747",
    pages = "8440--8451"
}

@inproceedings{probing-classifiers-under-the-hood,
    title = "Under the Hood: Using Diagnostic Classifiers to Investigate and Improve how Language Models Track Agreement Information",
    author = "Giulianelli, Mario  and
      Harding, Jack  and
      Mohnert, Florian  and
      Hupkes, Dieuwke  and
      Zuidema, Willem",
    editor = "Linzen, Tal  and
      Chrupa{\l}a, Grzegorz  and
      Alishahi, Afra",
    booktitle = "Proceedings of the 2018 {EMNLP} Workshop {B}lackbox{NLP}: Analyzing and Interpreting Neural Networks for {NLP}",
    month = nov,
    year = "2018",
    address = "Brussels, Belgium",
    publisher = "Association for Computational Linguistics",
    url = "https://aclanthology.org/W18-5426/",
    doi = "10.18653/v1/W18-5426",
    pages = "240--248"
}

@article{cola,
    title = "Neural Network Acceptability Judgments",
    author = "Warstadt, Alex  and
      Singh, Amanpreet  and
      Bowman, Samuel R.",
    editor = "Lee, Lillian  and
      Johnson, Mark  and
      Roark, Brian  and
      Nenkova, Ani",
    journal = "Transactions of the Association for Computational Linguistics",
    volume = "7",
    year = "2019",
    address = "Cambridge, MA",
    publisher = "MIT Press",
    url = "https://aclanthology.org/Q19-1040/",
    doi = "10.1162/tacl_a_00290",
    pages = "625--641"
}

@misc{turblimp,
author = {Başar, Ezgi and Padovani, F.P and Jumelet, Jaap and Bisazza, Arianna},
year = {2025},
month = {06},
pages = {},
title = {TurBLiMP: A Turkish Benchmark of Linguistic Minimal Pairs},
doi = {10.48550/arXiv.2506.13487}
}

@misc{german-blimp,
      title={Evaluating German Transformer Language Models with Syntactic Agreement Tests}, 
      author={Karolina Zaczynska and Nils Feldhus and Robert Schwarzenberg and Aleksandra Gabryszak and Sebastian Möller},
      year={2020},
      eprint={2007.03765},
      archivePrefix={arXiv},
      primaryClass={cs.CL},
      url={https://arxiv.org/abs/2007.03765}, 
}

@inproceedings{russian-blimp,
	location = {Miami, Florida, {USA}},
	title = {{RuBLiMP}: Russian Benchmark of Linguistic Minimal Pairs},
        year= {2024},
	url = {https://aclanthology.org/2024.emnlp-main.522},
	doi = {10.18653/v1/2024.emnlp-main.522},
	shorttitle = {{RuBLiMP}},
	eventtitle = {Proceedings of the 2024 Conference on Empirical Methods in Natural Language Processing},
	pages = {9268--9299},
	booktitle = {Proceedings of the 2024 Conference on Empirical Methods in Natural Language Processing},
	publisher = {Association for Computational Linguistics},
	author = {Taktasheva, Ekaterina and Bazhukov, Maxim and Koncha, Kirill and Fenogenova, Alena and Artemova, Ekaterina and Mikhailov, Vladislav},
	urldate = {2025-06-24},
	date = {2024},
	langid = {english},
}

@inproceedings{spanish-blimp,
    title = "{E}s{C}o{LA}: {S}panish Corpus of Linguistic Acceptability",
    author = "Bel, N{\'u}ria  and
      Punsola, Marta  and
      Ruiz-Fern{\'a}ndez, Valle",
    editor = "Calzolari, Nicoletta  and
      Kan, Min-Yen  and
      Hoste, Veronique  and
      Lenci, Alessandro  and
      Sakti, Sakriani  and
      Xue, Nianwen",
    booktitle = "Proceedings of the 2024 Joint International Conference on Computational Linguistics, Language Resources and Evaluation (LREC-COLING 2024)",
    month = may,
    year = "2024",
    address = "Torino, Italia",
    publisher = "ELRA and ICCL",
    url = "https://aclanthology.org/2024.lrec-main.554/",
    pages = "6268--6277"
}

@inproceedings{icelandic-and-other-langs-blimp,
    title = "{MELA}: Multilingual Evaluation of Linguistic Acceptability",
    author = "Zhang, Ziyin  and
      Liu, Yikang  and
      Huang, Weifang  and
      Mao, Junyu  and
      Wang, Rui  and
      Hu, Hai",
    editor = "Ku, Lun-Wei  and
      Martins, Andre  and
      Srikumar, Vivek",
    booktitle = "Proceedings of the 62nd Annual Meeting of the Association for Computational Linguistics (Volume 1: Long Papers)",
    month = aug,
    year = "2024",
    address = "Bangkok, Thailand",
    publisher = "Association for Computational Linguistics",
    url = "https://aclanthology.org/2024.acl-long.146/",
    doi = "10.18653/v1/2024.acl-long.146",
    pages = "2658--2674"
}

@inproceedings{chinese-blimp,
    title = "{SLING}: {S}ino Linguistic Evaluation of Large Language Models",
    author = "Song, Yixiao  and
      Krishna, Kalpesh  and
      Bhatt, Rajesh  and
      Iyyer, Mohit",
    editor = "Goldberg, Yoav  and
      Kozareva, Zornitsa  and
      Zhang, Yue",
    booktitle = "Proceedings of the 2022 Conference on Empirical Methods in Natural Language Processing",
    month = dec,
    year = "2022",
    address = "Abu Dhabi, United Arab Emirates",
    publisher = "Association for Computational Linguistics",
    url = "https://aclanthology.org/2022.emnlp-main.305/",
    doi = "10.18653/v1/2022.emnlp-main.305",
    pages = "4606--4634"
}

@inproceedings{japanese-blimp,
    title = "{JBL}i{MP}: {J}apanese Benchmark of Linguistic Minimal Pairs",
    author = "Someya, Taiga  and
      Oseki, Yohei",
    editor = "Vlachos, Andreas  and
      Augenstein, Isabelle",
    booktitle = "Findings of the Association for Computational Linguistics: EACL 2023",
    month = may,
    year = "2023",
    address = "Dubrovnik, Croatia",
    publisher = "Association for Computational Linguistics",
    url = "https://aclanthology.org/2023.findings-eacl.117/",
    doi = "10.18653/v1/2023.findings-eacl.117",
    pages = "1581--1594"
}

@inproceedings{someya-etal-2024-targeted,
    title = "Targeted Syntactic Evaluation on the {C}homsky Hierarchy",
    author = "Someya, Taiga  and
      Yoshida, Ryo  and
      Oseki, Yohei",
    editor = "Calzolari, Nicoletta  and
      Kan, Min-Yen  and
      Hoste, Veronique  and
      Lenci, Alessandro  and
      Sakti, Sakriani  and
      Xue, Nianwen",
    booktitle = "Proceedings of the 2024 Joint International Conference on Computational Linguistics, Language Resources and Evaluation (LREC-COLING 2024)",
    month = may,
    year = "2024",
    address = "Torino, Italia",
    publisher = "ELRA and ICCL",
    url = "https://aclanthology.org/2024.lrec-main.1356/",
    pages = "15595--15605"
}

@inproceedings{newman-2021-refining-tse,
    title = "Refining Targeted Syntactic Evaluation of Language Models",
    author = "Newman, Benjamin  and
      Ang, Kai-Siang  and
      Gong, Julia  and
      Hewitt, John",
    editor = "Toutanova, Kristina  and
      Rumshisky, Anna  and
      Zettlemoyer, Luke  and
      Hakkani-Tur, Dilek  and
      Beltagy, Iz  and
      Bethard, Steven  and
      Cotterell, Ryan  and
      Chakraborty, Tanmoy  and
      Zhou, Yichao",
    booktitle = "Proceedings of the 2021 Conference of the North American Chapter of the Association for Computational Linguistics: Human Language Technologies",
    month = jun,
    year = "2021",
    address = "Online",
    publisher = "Association for Computational Linguistics",
    url = "https://aclanthology.org/2021.naacl-main.290/",
    doi = "10.18653/v1/2021.naacl-main.290",
    pages = "3710--3723"
}

@inproceedings{evaluation-syntactic-knowledge-low-res,
    title = "Controlled Evaluation of Syntactic Knowledge in Multilingual Language Models",
    author = "Kryvosheieva, Daria  and
      Levy, Roger",
    editor = "Hettiarachchi, Hansi  and
      Ranasinghe, Tharindu  and
      Rayson, Paul  and
      Mitkov, Ruslan  and
      Gaber, Mohamed  and
      Premasiri, Damith  and
      Tan, Fiona Anting  and
      Uyangodage, Lasitha",
    booktitle = "Proceedings of the First Workshop on Language Models for Low-Resource Languages",
    month = jan,
    year = "2025",
    address = "Abu Dhabi, United Arab Emirates",
    publisher = "Association for Computational Linguistics",
    url = "https://aclanthology.org/2025.loreslm-1.30/",
    pages = "402--413"
}

@ARTICLE{kulmizev-schrodinger-2021,
AUTHOR={Kulmizev, Artur  and Nivre, Joakim },
TITLE={Schrödinger's tree—On syntax and neural language models},      
JOURNAL={Frontiers in Artificial Intelligence},
VOLUME={Volume 5 - 2022},
YEAR={2022},
URL={https://www.frontiersin.org/journals/artificial-intelligence/articles/10.3389/frai.2022.796788},
DOI={10.3389/frai.2022.796788},
ISSN={2624-8212}}

@misc{dutch-blimp,
    author = {Suijkerbuijk, Michelle and Prins, Zoë and Kloots, Marianne de Heer and Zuidema, Willem and Frank, Stefan L.},
    title = {BLiMP-NL: A Corpus of Dutch Minimal Pairs and Acceptability Judgments for Language Model Evaluation},
    journal = {Computational Linguistics},
    pages = {1-35},
    year = {2025},
    month = {05},
    issn = {0891-2017},
    doi = {10.1162/coli_a_00559},
    url = {https://doi.org/10.1162/coli_a_00559},
    eprint = {https://direct.mit.edu/coli/article-pdf/doi/10.1162/coli_a_00559/2512113/coli_a_00559.pdf},
}

@inproceedings{chinese-blimp-test,
    title = "Controlled Evaluation of Grammatical Knowledge in {M}andarin {C}hinese Language Models",
    author = "Wang, Yiwen  and
      Hu, Jennifer  and
      Levy, Roger  and
      Qian, Peng",
    editor = "Moens, Marie-Francine  and
      Huang, Xuanjing  and
      Specia, Lucia  and
      Yih, Scott Wen-tau",
    booktitle = "Proceedings of the 2021 Conference on Empirical Methods in Natural Language Processing",
    month = nov,
    year = "2021",
    address = "Online and Punta Cana, Dominican Republic",
    publisher = "Association for Computational Linguistics",
    url = "https://aclanthology.org/2021.emnlp-main.454/",
    doi = "10.18653/v1/2021.emnlp-main.454",
    pages = "5604--5620"
}

@inproceedings{swedish-blimp,
    title = "{D}a{LAJ} {--} a dataset for linguistic acceptability judgments for {S}wedish",
    author = "Volodina, Elena  and
      Mohammed, Yousuf Ali  and
      Klezl, Julia",
    editor = {Alfter, David  and
      Volodina, Elena  and
      Pilan, Ildik{\'o}  and
      Gra{\"e}n, Johannes  and
      Borin, Lars},
    booktitle = "Proceedings of the 10th Workshop on NLP for Computer Assisted Language Learning",
    month = may,
    year = "2021",
    address = "Online",
    publisher = "LiU Electronic Press",
    url = "https://aclanthology.org/2021.nlp4call-1.3/",
    pages = "28--37"
}

@misc{irish-blimp,
      title={Irish-BLiMP: A Linguistic Benchmark for Evaluating Human and Language Model Performance in a Low-Resource Setting}, 
      author={Josh McGiff and Khanh-Tung Tran and William Mulcahy and Dáibhidh Ó Luinín and Jake Dalzell and Róisín Ní Bhroin and Adam Burke and Barry O'Sullivan and Hoang D. Nguyen and Nikola S. Nikolov},
      year={2025},
      eprint={2510.20957},
      archivePrefix={arXiv},
      primaryClass={cs.CL},
      url={https://arxiv.org/abs/2510.20957}, 
}

@misc{multi-blimp,
      title={MultiBLiMP 1.0: A Massively Multilingual Benchmark of Linguistic Minimal Pairs}, 
      author={Jaap Jumelet and Leonie Weissweiler and Joakim Nivre and Arianna Bisazza},
      year={2025},
      eprint={2504.02768},
      archivePrefix={arXiv},
      primaryClass={cs.CL},
      url={https://arxiv.org/abs/2504.02768}, 
}

\end{document}